\documentclass[10pt,twocolumn,letterpaper]{article}

\usepackage{iccv}
\usepackage{times}
\usepackage{epsfig}
\usepackage{graphicx}
\usepackage{amsmath}
\usepackage{amssymb}

\usepackage[ruled,vlined]{algorithm2e}
\usepackage{subcaption}
\usepackage{array}
\usepackage{siunitx}
\newcommand*{\affaddr}[1]{#1} 
\newcommand*{\affmark}[1][*]{\textsuperscript{#1}}
\newcommand*{\email}[1]{\texttt{#1}}
\usepackage[breaklinks=true,bookmarks=false]{hyperref}

\iccvfinalcopy % *** Uncomment this line for the final submission

\def\iccvPaperID{4192} % *** Enter the ICCV Paper ID here

\ificcvfinal\fi

\begin{document}

\newcolumntype{L}[1]{>{\raggedright\let\newline\\\arraybackslash\hspace{0pt}}m{#1}}
\newcolumntype{C}[1]{>{\centering\let\newline\\\arraybackslash\hspace{0pt}}m{#1}}
\newcolumntype{R}[1]{>{\raggedleft\let\newline\\\arraybackslash\hspace{0pt}}m{#1}}

%%%%%%%%% TITLE
\title{AutoGAN: Neural Architecture Search for Generative Adversarial Networks}

\author{%
Xinyu Gong\affmark[1] \quad Shiyu Chang\affmark[2] \quad Yifan Jiang\affmark[1] \quad Zhangyang Wang\affmark[1]\\
\affaddr{\affmark[1]Department of Computer Science \& Engineering, Texas A\&M University \\ \affmark[2]MIT-IBM Watson AI Lab}
%\affaddr{}
\\
\email{\{xy\_gong, yifanjiang97, atlaswang\}@tamu.edu} \quad \email{shiyu.chang@ibm.com}\\
%\affaddr{\LaTeX\ University}%
}

\maketitle
\ificcvfinal\thispagestyle{empty}\fi

%%%%%%%%% ABSTRACT
\begin{abstract}
     Neural architecture search (NAS) has witnessed prevailing success in image classification and (very recently) segmentation tasks. In this paper, we present the first preliminary study on introducing the NAS algorithm to generative adversarial networks (GANs), dubbed \textbf{AutoGAN}. The marriage of NAS and GANs faces its unique challenges. We define the search space for the generator architectural variations and use an RNN controller to guide the search, with parameter sharing and dynamic-resetting to accelerate the process. Inception score is adopted as the reward, and a multi-level search strategy is introduced to perform NAS in a progressive way. Experiments validate the effectiveness of AutoGAN on the task of unconditional image generation. Specifically, our discovered architectures achieve highly competitive performance compared to current state-of-the-art hand-crafted GANs, e.g., setting new state-of-the-art FID scores of 12.42 on CIFAR-10, and 31.01 on STL-10, respectively. We also conclude with a discussion of the current limitations and future potential of AutoGAN. The code is avaliable at \url{https://github.com/TAMU-VITA/AutoGAN}.
\end{abstract}

%%%%%%%%% BODY TEXT
\section{Introduction \label{intro}}

Generative adversarial networks (GANs) have been prevailing since its origin \cite{goodfellow2014generative}. One of their most notable successes lies in generating realistic natural images with various convolutional architectures \cite{radford2015unsupervised,brock2018large,xu2017attngan,karras2017progressive,gulrajani2017improved,zhang2019dada}. In order to improve the quality of generated images, many efforts have been proposed, including modifying discriminator loss functions \cite{arjovsky2017wasserstein,zhao2016energy}, enforcing regularization terms \cite{brock2016neural,gulrajani2017improved,miyato2018spectral,brock2018large}, introducing attention mechanism \cite{xu2017attngan,zhang2018self}, and adopting progressive training \cite{karras2017progressive}.

However, the backbone architecture design of GANs has received relatively less attention, and was often considered a less significant factor accounting for GAN performance \cite{lucic2018gans,kurach2018gan}. Most earlier GANs stick to relatively shallow generator and discriminator architectures, mostly owing to the notorious instability in GAN training.  Lately, several state-of-the-art GANs \cite{gulrajani2017improved,miyato2018spectral,zhang2018self,brock2018large} adopted deep residual network generator backbones for better generating high-resolution images. In most other computer vision tasks, a lot of the progress arises from the improved design of network architectures, such as image classification \cite{he2016deep, he2016identity,szegedy2016rethinking,szegedy2015going,lin2017feature}, segmentation \cite{chen2018deeplab, chen2017rethinking,ronneberger2015u}, and pose estimation \cite{newell2016stacked}. Hereby, we advocate that enhanced backbone designs are also important for improving GANs further.

In recent years, there are surging interests in designing sophisticated neural network architectures automatically. Neural architecture search (NAS) has been successfully developed and evaluated on the task of image classification \cite{zoph2016neural,Pham:tl}, and lately on image segmentation as well \cite{chen2018searching,liu2019auto}. The discovered architectures outperform human-designed models. However, naively porting existing NAS ideas from image classification/segmentation to GAN would not suffice. First and foremost, even given hand-designed architectures, the training of GANs is notoriously unstable and prone to collapse \cite{salimans2016improved}. Mingling NAS into the training process will undoubtedly amplify the difficulty. As another important challenge, while the validation accuracy makes a natural reward option for NAS in image classification, it is less straightforward to choose a good metric for evaluating and guiding the GAN search process.  

This paper presents an architecture search scheme specifically tailored for GANs, dubbed \textbf{AutoGAN}. Up to our best knowledge, AutoGAN describes \textbf{the first} attempt to incorporate NAS with GANs, and belongs to \textbf{one of the first} attempts to extend NAS \textit{beyond image classification} too. Our technical innovations are summarized as follows:
\begin{itemize}
    % \item Up to our best knowledge, AutoGAN describes the first attempt to incorporate NAS with GANs, and is one of the first attempts to extend NAS beyond image classification too. 
     \vspace{-0.3em}
    \item We define the search space to capture the GAN architectural variations. On top of that, we use an RNN controller \cite{zoph2016neural} to guide the architecture search. Based on the parameter sharing strategy in \cite{Pham:tl}, we further introduce a parameter dynamic-resetting strategy in our search progress, to boost the training speed. 
         \vspace{-0.3em}
    \item We use Inception score (IS) \cite{salimans2016improved} as the reward, in the reinforcement-learning-based optimization of AutoGAN. The discovered models are found also to show favorable performance under other GAN metrics, e.g., the Fréchet Inception Distance (FID) \cite{salimans2016improved}.
         \vspace{-0.3em}
    \item We further introduce multi-level architecture search (MLAS) to AutoGAN, as motivated by the progressive GAN training \cite{karras2017progressive}. MLAS performs the search in multiple stages, in a bottom-up sequential fashion, with beam search \cite{Liu:2018tr}. 
\vspace{-0.3em}
\end{itemize}
We conduct a variety of experiments to validate the effectiveness of AutoGAN. Our discovered architectures yield highly promising results that are better than or comparable with current hand-designed GANs. On the CIFAR-10 dataset, AutoGAN obtains an Inception score of \textbf{8.55}, and a FID score of \textbf{12.42}. In addition, we show that the discovered architecture on CIFAR-10 even performs competitively on the STL-10 image generation task, with a \textbf{9.16} Inception score and a \textbf{31.01} FID score, demonstrating a strong transferability. On both datasets, AutoGAN establishes the new state-of-the-art FID scores. Many of our experimental findings concur with previous GAN crafting experiences, and shed light on new insights into GAN generator design. 

%-------------------------------------------------------------------------
\section{Related Work}
\subsection{Neural Architecture Search \label{nas}}
Neural architecture search (NAS) algorithms aim to find an optimal neural network architecture instead of using a hand-crafted one for a specific task. Previous works on NAS have achieved great success on the task of image classification \cite{krizhevsky2012imagenet}. Recent works further extended the NAS algorithms to dense and structured prediction \cite{chen2018searching,liu2019auto}. It is also worth mentioning that NAS is also applied to CNN compression and acceleration \cite{he2018amc}. However, there is not any NAS algorithm developed for generative models. 

A NAS algorithm consists of three key components: the search space, the optimization algorithm, and the proxy task. For search space, there are generally two categories: searching for the whole architecture directly (macro search), or searching for cells and stacking them in a pre-defined way (micro search). For the optimization algorithm, popular options include reinforcement learning \cite{baker2016designing,zoph2016neural,zhong2018practical,Zoph:2018ta}, evolutionary algorithm \cite{xie2017genetic}, Bayesian optimization \cite{jin2018efficient}, random search \cite{chen2018searching}, and gradient-based optimization methods \cite{Liu:th,Ahmed:2018fc}. For the proxy task, it is designed to evaluate the performance of discovered architecture efficiently during training. Examples include early stop \cite{chen2018searching, Zoph:2018ta}, using low-resolution images \cite{wang2016studying, chrabaszcz2017downsampled, krizhevsky2009learning}, performance prediction with a surrogate model \cite{Liu:2018tr}, employing a small backbone \cite{chen2018searching} or leveraging shared parameters \cite{Pham:tl}. 

Most NAS algorithms \cite{zoph2016neural,Pham:tl} generate the network architecture (macro search) or the cell (micro search) by one pass of the controller. A recent work \cite{Liu:2018tr} introduced multi-level search to NAS in the image classification task, using beam search. Architecture search will begin on a smaller cell, and top-performance candidates are preserved. The next round of search will be continued based on them for the bigger cell.

\subsection{Generative Adversarial Network}
%Generative adversarial networks (GANs) are one of the most appealing generative models currently, which is first presented by Goodfellow \etal \cite{goodfellow2014generative}. 
A GAN has a generator network and a discriminator network playing a min-max two-player game against each other. It has achieved great success in many generation and synthesis tasks, such as text-to-image translation \cite{zhang2017stackgan,zhang2017stackgan++,xu2017attngan,Reed}, image-to-image translation \cite{isola2017image,Zhu:2017ua,yang2019controllable}, and image enhancement \cite{ledig2017photo, kupyn2018deblurgan,jiang2019enlightengan}. However, the training of GAN is often found to be highly unstable \cite{salimans2016improved}, and commonly suffers from non-convergence, mode collapse, and sensitivity to hyperparameters. Many efforts have been devoted to alleviating those problems, such as the Wasserstein loss \cite{arjovsky2017wasserstein}, spectral normalization \cite{miyato2018spectral}, progressive training \cite{karras2017progressive}, and the self-attention block \cite{zhang2018self}, to name just a few. 

\section{Technical Approach}

A GAN consists of two competing networks: a generator and a discriminator. It is also known that the two architectures have to be delicately balanced in their learning capacities.
%such that one will not easily overwhelm the other and results in trivial learning. 
Hence to build AutoGAN, the first question is: \textit{how to build the two networks in a GAN} (generator and discriminator, denoted as $G$ and $D$ hereinafter) together? On one hand, if we use a pre-fixed $D$ (or $G$) and search only for $G$ (or $D$), it will easily incur imbalance between the powers of $D$ or $G$ \cite{heusel2017gans,arjovsky2017towards} (especially, at the early stage of NAS), resulting in slow updates or trivial learning. On the other hand, while it might be possible jointly search for $G$ and $D$, empirical experiments observe that such two-way NAS will further deteriorate the original unstable GAN training, leading to highly oscillating training curves and often failure of convergence. As a trade-off, we propose to use NAS to \textbf{only search for the architecture of $G$}, while growing $D$ as $G$ becomes deeper, by following a given routine to stack pre-defined blocks. The details of growing $D$ will be explained more in the supplementary. 

\begin{figure*}[h]
\begin{center}
 \includegraphics[width=\linewidth]{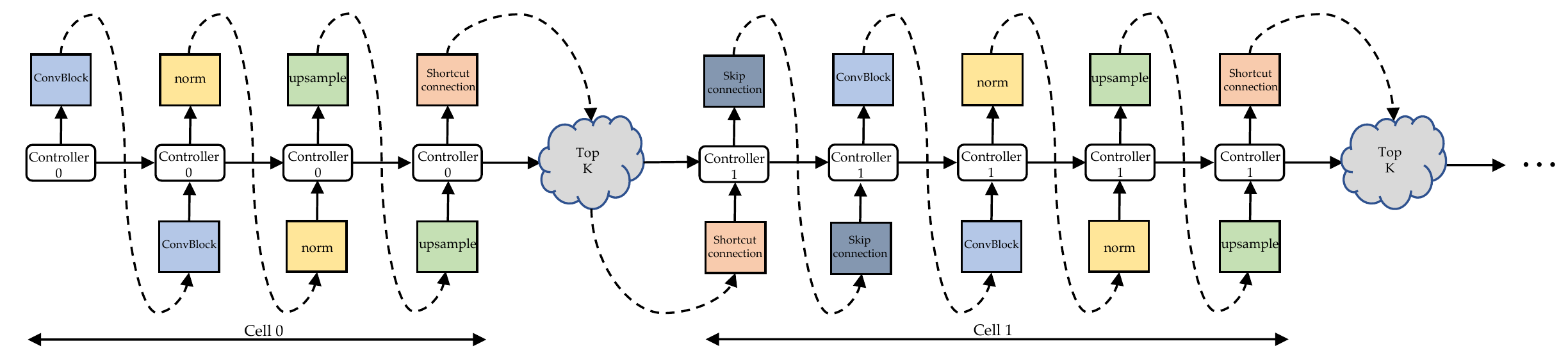}
\end{center}
\vspace{-1em}
   \caption{The running scheme of the RNN controller. At each time step, the controller outputs a hidden vector to be decoded into an operation, with its corresponding softmax classifier. Note that, using MLAS, we use a different controller for the search of each cell. Once the search process for one cell is done, the controller samples $M$ candidate architectures and then picks the top $K$ from them. The top $K$ architectures' controller outputs will fed as the input of the next cell's controller.}
   \vspace{-1em}
\label{fig:controller}
\end{figure*}

Based on that, AutoGAN follows the basic idea of \cite{zoph2016neural} to use a recurrent neural network (RNN) controller to choose blocks from its search space, to build the $G$ network. The basic scheme is illustrated in Figure \ref{fig:controller}. We make multi-fold innovations to address the unique challenges arising from the specific task of training GANs. We next introduce AutoGAN from the three key aspects: the search space, the proxy task and the optimization algorithm. 

\subsection{Search Space}

\begin{figure}[h]
\begin{center}
 \includegraphics[width=0.9\linewidth]{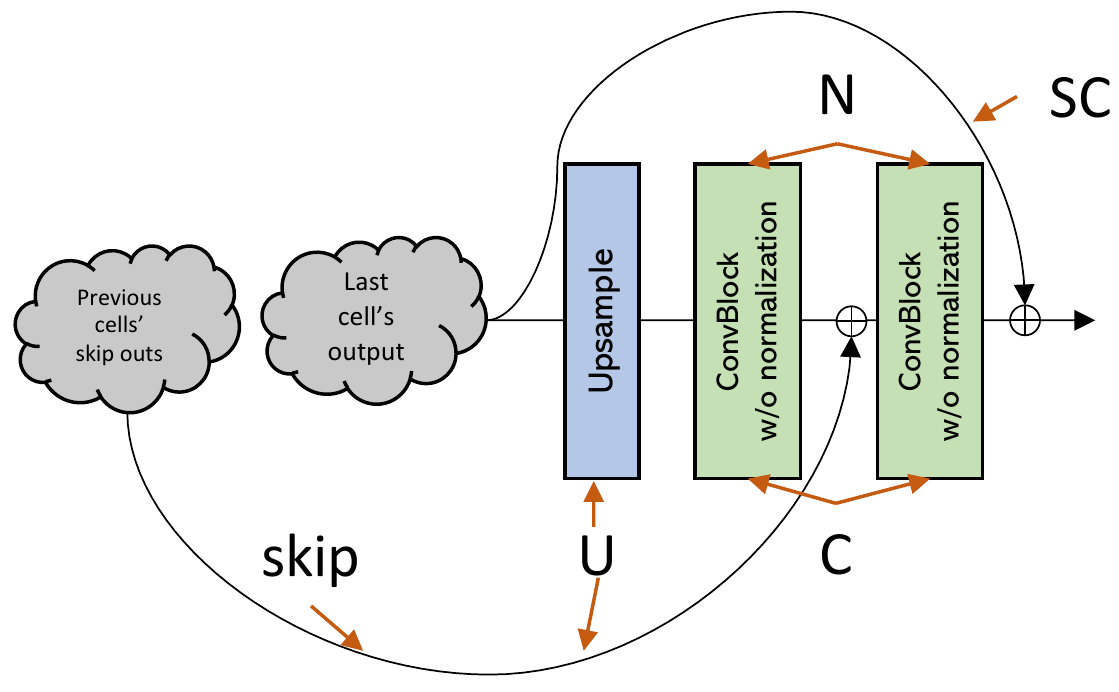}
\end{center}
\vspace{-1em}
   \caption{The search space of a generator cell in AutoGAN.}
   \vspace{-1em}
\label{fig:ss}
\end{figure}

% AutoGAN is based on the macro search strategy, where specific operations will be selected layer by layer. Thus, the whole search space is parameterized by the total number of searched layers and the operation types of each layer. 
AutoGAN is based on a multi-level architecture search strategy, where the generator is composed of several cells. Here we use a \textbf{(s+5)} \textbf{element tuple} $(skip_1, ..., skip_s, C, N, U, SC)$ to categorize the $s$-th cell, where $s$ is the cell index starting at 0 (0-th cell doesn't have $skip_0$ connection):
\begin{itemize}
    \vspace{-0.3em}
    \item $skip_i$ is a binary value indicating whether the current $s$-th cell takes a skip connection from the $(i-1)$-th cell as its input, $i = 1, ..., s$. Note that each cell could take multiple skip connections from other preceding cells. 
    \item $C$ is the basic convolution block type, including pre-activation \cite{he2016identity} and post-activation convolution block.\vspace{-0.3em}
    \item $N$ stands for the normalization type of this block, including three options: batch normalization\cite{ioffe2015batch}, instance normalization\cite{ulyanov2016instance}, and no normalization. \vspace{-0.3em}
    \item $U$ stands for the upsampling operation which was standard in current image generation GANs, including bilinear upsampling, nearest neighbour upsampling, and stride 2 deconvolution. 
    \vspace{-0.3em}
    \item $SC$ is a binary value indicating the in-cell shortcut.
\end{itemize} 
Fig. \ref{fig:ss} illustrates the AutoGAN generator search space. The upsampling operation $U$ will also determine the upsample method of the skip-in feature map.

\vspace{-1pt}

\subsection{Proxy Task}

Inception score  (IS) \cite{salimans2016improved} and FID score \cite{heusel2017gans} are two main evaluation metrics for GANs. Since FID score is much more time-consuming to calculate, we choose the IS of each derived child model, as the reward to update controller via reinforcement learning. 

Parameter sharing \cite{luong2015multi,zoph2016transfer} shows to effectively boost the efficiency of NAS \cite{Pham:tl}. Based on the parameter-sharing in \cite{Pham:tl}, we further bring in a parameter \textbf{dynamic-resetting} strategy to AutoGAN. It has been observed that the training of GANs becomes unstable and undergoes mode collapse after long time training \cite{brock2018large}. It could be a waste of time to continue training the shared collapsed model. Empirically, we observe the variance of the training loss (hinge loss) usually becomes very small when mode collapse happens.

Based on this observation, we set a moving window to store the most recent training loss values, for both generator and discriminator. Once the standard deviation of those stored training losses is smaller than a pre-defined threshold, the training of the shared GAN of current iteration will be terminated.
%and turned to train controller. 
The parameters of the shared-GAN model will be re-initialized after updating the controller at the current iteration. Notice that we do \textbf{NOT} re-initialize the parameters of the RNN controller, thus it could continue to guide the architecture search with inheriting the historic knowledge. With dynamic-resetting, the searching process becomes much more efficient. %, and we observe it to help bypass potential mode-collapses. 

%as we skip the training on potential mode-collapsed shared GAN model.

\subsection{Optimization Method}
There are two sets of parameters in the AutoGAN: the RNN controller's parameters (denoted as $\theta$); and the shared GAN parameters in the searched generator and the corresponding discriminator (denoted as $\omega$). The training process is briefly outlined in Algo. \ref{algo1}, as an alternating process between two phases.  

The first phase will train $\omega$ of the shared GAN for several epochs, with $\theta$ fixed. For each training iteration of the shared GAN training process, a candidate architecture will be sampled by the RNN controller. Once the standard deviation of the recorded training loss drop below the threshold, the dynamic resetting FLAG $F_{DR}$ will be set to $True$ and the training process of shared GAN will be terminated immediately. Note that the shared GAN will not be re-initialized until the training of controller at the current epoch completes. The second phase trains $\theta$ with $\omega$ fixed: the controller will first sample $K$ child models of the shared generator. Their ISs will be calculated as rewards. The RNN controller will be updated via reinforcement learning with a moving average baseline. After training $u_{stage}$ iterations, the top $K$ architectures will be picked from the derived architectures. Meanwhile, a new controller will be initialized to proceed the architecture searching of next stage.

\begin{algorithm}
\SetAlgoLined
\SetKwData{U}{$u_{stage}$}
\SetKwData{ITERS}{iters}
\SetKwData{STAGE}{stage}
\SetKwFunction{TRAIN}{train}
\SetKwFunction{NEW}{new}
\SetKwFunction{GROW}{grow}
\SetKwFunction{INITIALIZE}{initialize}

$\ITERS = 0$ \;
$\STAGE = 0$ \;
$F_{DR} = False$ \;
\While{ $\ITERS < 90$ }{
    \TRAIN($generator$, $discriminator$, $F_{DR}$)\; 
    \TRAIN($controller$)\;
    \If{$\ITERS\quad \% \quad \U == 0 $}{
       save the top $K$ architectures\;
       $generator$ = \GROW($generator$) \;
       $discriminator$ = \GROW($discriminator$) \;
       $controller$ = \NEW($controller$)\;
       $\STAGE+=1$\;
       } 
    \If{$F_{DR} == True$}{
        \tcp{dynamic reset}
        \INITIALIZE($generator$)\;
        \INITIALIZE($discriminator$)\;
        $F_{DR} =False$\;
       }
    $\ITERS += 1$\;
}
 \caption{Pseudo codes for AutoGAN searching.}\label{algo1}
\end{algorithm}
\vspace{-1em}

\subsubsection{Training Shared GAN}
During this phase, we fix the RNN controller's policy $\pi(\textbf{a}, \theta)$ and update the shared parameters $\omega$ through standard GAN training. Specifically, we train with an alternating fashion using the hinge adversarial loss \cite{miyato2018spectral,brock2018large,tran2017deep,zhang2018self}:
\begin{equation}
	\begin{split}
    \mathcal{L}_D = & E_{x\sim q_{data}}[min(0, -1 + D(x)] 
    + \\& E_{z\sim p(z)}[min(0, -1-D(G(z))],
    \end{split}
\end{equation}
\begin{equation}
    \mathcal{L}_G = E_{z\sim p(z)}[min(0, D(G(z))],
\end{equation}

We further introduce multi-level architecture search (\textbf{MLAS}) to AutoGAN, where the generator (and correspondingly the discriminator) will grow progressively. MLAS performs the search in a bottom-up cell-wise fashion, using beam search \cite{Liu:2018tr}. When searching for the next cell, we will use a different controller, selects the top $K$ beams from current candidate cells and start the next round of search based on them. 

\subsubsection{Training the Controller}
 In this phase we fix $\omega$ and update the policy parameters $\theta$ of the controller. We define the reward $\mathcal{R}(a, \omega)$ as the IS of the sampled child model $a$. The RNN controller is updated using the Adam optimizer \cite{kingma2014adam} via REINFORCE \cite{williams1992simple}, with a moving average baseline. Besides, we also add a entropy term to encourage the exploration.
 
We use a LSTM \cite{hochreiter1997long} controller. For each time step, the LSTM will output a hidden vector, which will be decoded and classified by its corresponding softmax classifier. The LSTM controller works in an autogressive way, where the output of the last step will be fed into the next step. The operations of GAN's each cell will be sampled from each time step's output. Specifically, a new controller will be initialized when a new cell is added to the existing model to increase the output image resolution. The previous top $K$ models' architectures and corresponding hidden vectors will be saved. Their hidden vectors will be fed into the new controller as input to search for next cell's operations.

\subsubsection{Architecture Derivation}
We will first sample several generator architectures from the learned policy $\pi(\textbf{a}, \theta)$. Then, the reward $R$ (Inception score) will be calculated for each model. We will then pick top $K$ models in terms of highest rewards, and train them from scratch. After that, we evaluate their Inception scores again, and the model with the highest Inception score becomes our final derived generator architecture. 

% \subsubsection{Designing Convolutional Neural Networks}
% Here we carfully discuss 

\section{Experiments}
\paragraph{Datasets}
In this paper, we adopt CIFAR-10 \cite{krizhevsky2009learning} as the main testbed for AutoGAN. It consists of 50,000 training image and 10,000 test images, where each image is of $32 \times 32$ resolution. We use the training set to train AutoGAN, without any data augmentation. 

We also adopt the STL-10 dataset to show the transferablity of AutoGAN discovered architectures. When using STL-10 for training, we adopt both the 5,000 image training set and 100,000 image unlabeled set. All images are resized to $48 \times 48$, without any other data augmentation.

\paragraph{Training details}

We follow the training setting of spectral normalization GAN \cite{miyato2018spectral} when training the shared GAN. The learning rate of both generator and discriminator are set to \num{2e-4}, using the hinge loss, an Adam optimizer \cite{kingma2014adam}, a batch size of 64 for discriminator and a batch size of 128 for generator. The spectral normalization is only enforced on the discriminator. We train our controller using Adam \cite{kingma2014adam}, with learning rate of \num{3.5e-4}. We also add the entropy of the controller's output probability to the reward, weighted by \num{1e-4}, in order to encourage the exploration.

AutoGAN is searched for 90 iterations. For each iteration, the shared GAN will be trained for 15 epochs, and the controller will be trained for 30 steps. We set the dynamic-resetting variance threshold at \num{1e-3}. We train the discovered architectures, using the same training setting as the shared GAN, for 50,000 generator iterations. %Meanwhile, the channel number of both generator and discriminator will be doubled.

\subsection{Results on CIFAR 10}

\begin{figure}[h]
\begin{center}
 \includegraphics[width=0.9\linewidth]{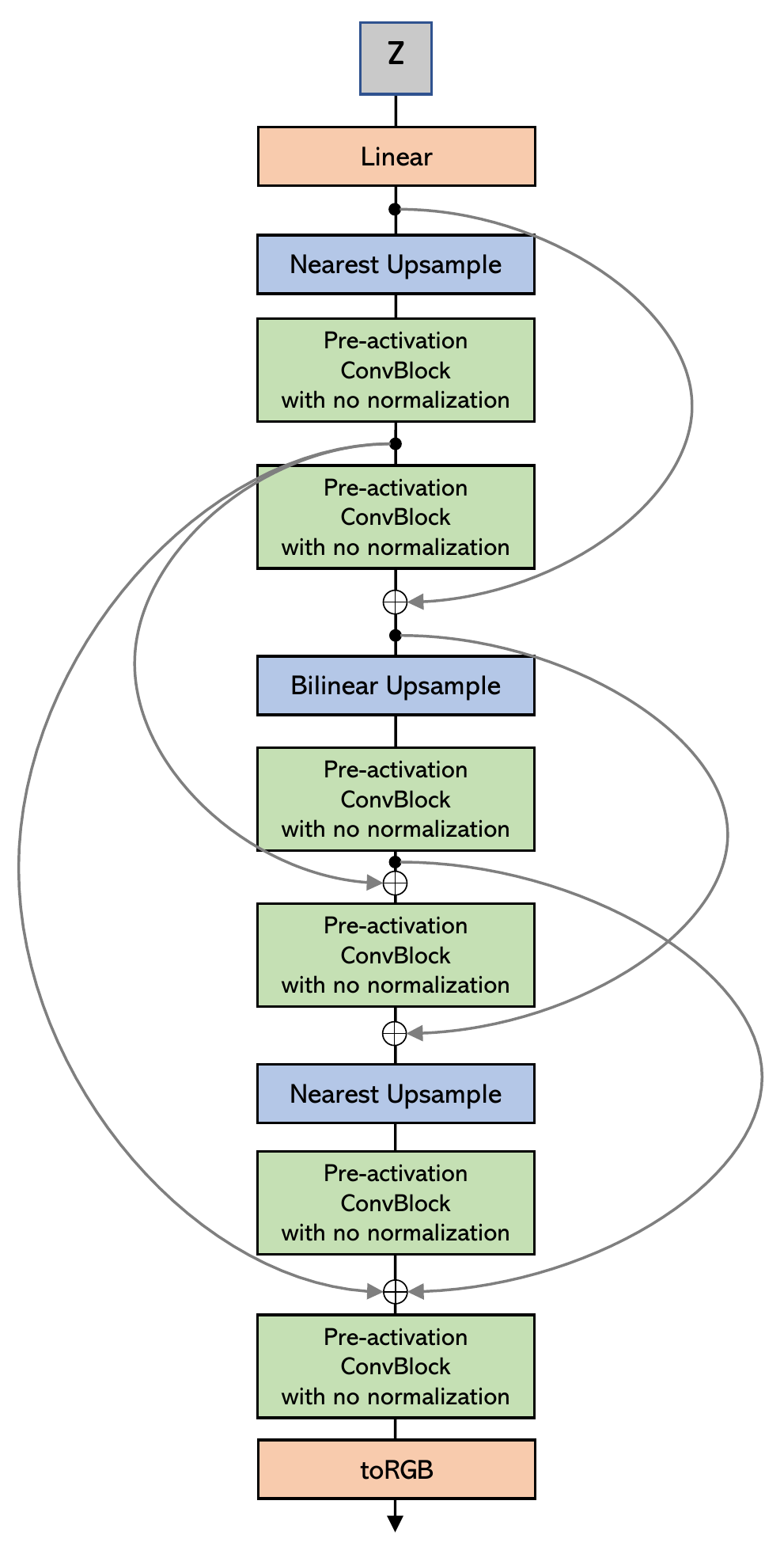}
\end{center}
\vspace{-1em}
   \caption{The AutoGAN (generator) architecture discovered by AutoGAN on CIFAR-10.}
\label{fig:result}
\end{figure}
\vspace{-1pt}

The generator architecture discovered by AutoGAN on the CIFAR-10 training set is displayed in Fig. \ref{fig:result}. For the task of unconditional CIFAR-10 image generation (no class labels used), \textbf{a number of notable observations could be summarized:}%
\begin{itemize}
    \vspace{-0.25em}
    \item The discovered architecture has 3 convolution blocks. AutoGAN clearly prefers pre-activation convolution block, than post-activation convolution block.  
    \vspace{-0.25em}
    \item AutoGAN highly prefers nearest neighbour upsample (also bilinear upsample) to deconvolution. That seems to coincide with previous experience \cite{odena2016deconvolution} that deconvolution might give rise to checkboard artifacts and nearest neighbour might thus be more favored. 
    \vspace{-0.25em}
    \item Interestingly enough, AutoGAN seems to prefer not using any normalization.
    \vspace{-0.25em}
    \item AutoGAN is clearly in favor of (denser) skipping connections. Especially, it is able to discover medium and long skips (skipping 2 - 4 convolutional layers), achieving multi-scale feature fusion. 
\end{itemize}

\begin{figure}[h]
\begin{center}
 \includegraphics[width=0.9\linewidth]{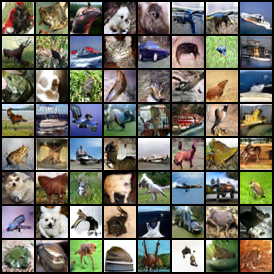}
\vspace{-1em}
\end{center}
   \caption{The generated CIFAR-10 results of AutoGAN. They are randomly sampled rather than cherry-picked.}
\label{fig:gen_cifar}
\end{figure}

We compare AutoGAN with recently published results by hand-crafted GANs on the CIFAR-10 dataset, in Table \ref{tab:comparison}. All results are collected from those original papers: therefore, they are obtained under their hand-tuned, best possible training settings. 
%For AutoGAN, we trained the discovered model three times to average reported inception socre and FID.
In terms of inception score, AutoGAN is slightly next to Progressive GAN \cite{karras2017progressive}, and surpasses many latest strong competitors such as SN-GAN \cite{miyato2018spectral}, improving MMD-GAN \cite{wang2018improving}, Dist-GAN \cite{tran2018dist}, MGAN \cite{hoang2018mgan}, and WGAN-GP \cite{gulrajani2017improved}. In terms of FID, AutoGAN outperforms all current state-of-the-art models. The visual examples of CIFAR-10 generated results are shown in Fig. \ref{fig:gen_cifar}.
%Overall, AutoGAN performs on par with Progressive GAN and MMD-GAN on CIFAR-10.

Note that the current search space of AutoGAN can only cover SN-GAN, but not the others; for example, WGAN-GP used the Wasserstein loss (with gradient clipping), and MGAN adopts a multi-discriminator structure (while our discriminator is very plain and not searched). By using the same group of building blocks, AutoGAN is able to outperform the hand-crafted SN-GAN: that \textbf{fair comparison} presents direct evidence for both the importance of generator structure, and the effectiveness of our search algorithm. 

Although not equipped with any explicit model complexity regularization yet, AutoGAN demonstrates the expected model parameter efficiency from NAS. The top discovered architecture in Fig. \ref{fig:result} has 1.77G FLOPs, while its performance is clearly better than SN-GAN (1.69G FLOPs) and comparable to Progressive GAN (6.39G FLOPs).

\begin{table}[h]
\caption{Inception score and FID score of unconditional image generation task on CIFAR-10. We achieve a state-of-the-art FID score of 12.42}
\centering
\resizebox{\linewidth}{!}{
    \begin{tabular}{ l || c | c }
    \hline
    Method & Inception score  & FID \\ \hline 
    DCGAN\cite{radford2015unsupervised} & $6.64 \pm .14$  & - \\ \hline
    Improved GAN\cite{salimans2016improved} & $6.86 \pm .06$ & - \\ \hline
    LRGAN \cite{yang2017lr} &$7.17 \pm .17$ & - \\ \hline
    DFM\cite{warde2017improving} &$7.72 \pm .13$ & - \\ \hline
    ProbGAN\cite{he2019probgan} & 7.75 & 24.60 \\ \hline
    WGAN-GP, ResNet \cite{gulrajani2017improved} & $7.86 \pm .07$  & - \\ \hline 
    Splitting GAN \cite{grinblat2017class} & $7.90 \pm .09$ & - \\ \hline 
    SN-GAN \cite{miyato2018spectral}& $8.22 \pm .05$ & $21.7 \pm .01$ \\ \hline 
    MGAN\cite{hoang2018mgan} & $8.33 \pm .10$ & 26.7 \\ \hline
    Dist-GAN \cite{tran2018dist} & - & $17.61 \pm .30$ \\ \hline
    Progressive GAN \cite{karras2017progressive} & \textbf{8.80} $\pm$ \textbf{.05} & - \\ \hline 
    Improving MMD GAN \cite{wang2018improving} & 8.29 & 16.21 \\ \hline
        \hline 
    AutoGAN-top1 (Ours) &  $8.55 \pm .10$  & \textbf{12.42} \\  \hline 
    AutoGAN-top2 & $8.42 \pm .07$  & $13.67$  \\  \hline
    AutoGAN-top3 & $8.41 \pm .11$  & $13.87$  \\  \hline
    \end{tabular}
    }
\label{tab:comparison}
\end{table}

\begin{figure}[h]
\begin{center}
 \includegraphics[width=0.99\linewidth]{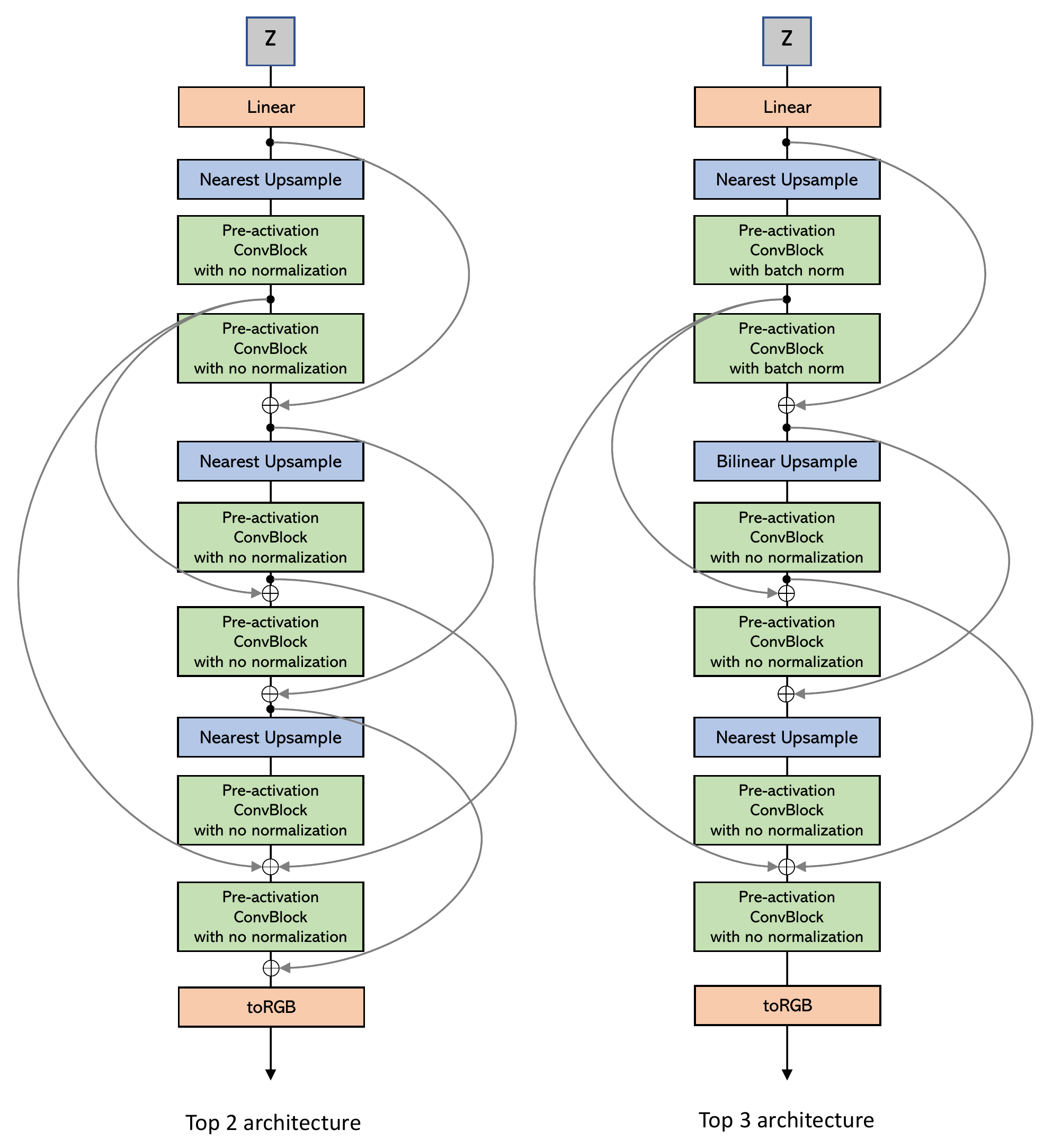}
\end{center}
\vspace{-1em}
   \caption{The top 2 and top 3 discovered architectures.}
   \vspace{-0.5em}
\label{fig:topk}
\end{figure}

% \vspace{-4pt}
We also report the Inception score and FID score of the candidate top 2 and top 3 discovered architectures in the bottom of Tab. \ref{tab:comparison}. Their corresponding architectures are shown in Fig. \ref{fig:topk}. We can see all of our top 3 discovered architecture achieves competitive performance with current state-of-the-art models.

% \vspace{-8pt}

\subsection{Transferability on STL-10}

The next question that we try to answer is: \textit{might the discovered architecture overfit the dataset}? In other words, would the same architecture stay powerful, when we use another dataset to re-train its weights (with the architecture structure fixed)? To address this question, we take the AutoGAN-discovered architecture on CIFAR-10, and re-train its weights on the STL-10 training and unlabled set, for the unconditional image generation task on STL-10. Note that STL-10 has higher image resolution than CIFAR-10. 

We compare with recently published results by hand-crafted GANs on the STL-10 dataset, in Table \ref{tab:stl}. Our result turns out to be surprisingly encouraging: while AutoGAN slightly lags behinds improving MMD-GAN in terms of inception score, it achieves the \textbf{best FID} result of 31.01 among all. The visual examples of STL-10 generated results are shown in Fig. 4. We expect to achieve even superior results on STL-10 if we re-perform the architecture search from the scratch, and leave it for future work. 

\begin{table}[h]
\caption{Inception score and FID score with unconditional image generation on STL-10. AutoGAN uses the discovered architecture on CIFAR-10.}
\label{tab:comparison}
\centering
\resizebox{\linewidth}{!}{
    \begin{tabular}{ l || c | c  }
    \hline
    Method & Inception score  & FID  \\ \hline 
    D2GAN \cite{nguyen2017dual} & $7.98$  & -  \\ \hline
    DFM \cite{warde2017improving} & $8.51 \pm .13$  & -  \\  \hline
    ProbGAN\cite{he2019probgan} & $8.87 \pm .095$ & $46.74$ \\ \hline
    SN-GAN \cite{miyato2018spectral}& $9.10 \pm .04$ & $40.1 \pm .50$  \\ \hline 
    Dist-GAN \cite{tran2018dist} & - & 36.19 \\ \hline
    Improving MMD GAN \cite{wang2018improving} & \textbf{9.34} & 37.63 \\ \hline
\hline
    AutoGAN (Ours) & $9.16 \pm .12$ & \textbf{31.01} \\  \hline
    \end{tabular}
    }
\label{tab:stl}
% \vspace{-8pt}
\end{table}
\begin{figure}[h]
\begin{center}
 \includegraphics[width=0.9\linewidth]{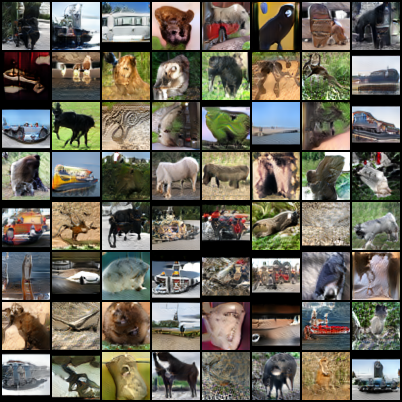}
\vspace{-1em}
\end{center}
   \caption{The generated STL-10 results of AutoGAN. They are randomly sampled rather than cherry-picked.}
\label{fig:gen_stl}
\vspace{-1em}
\end{figure}

%-------------------------------------------------------------------------
%-------------------------------------------------------------------------
%-------------------------------------------------------------------------
%-------------------------------------------------------------------------
\subsection{Ablation Study and Analysis}

\subsubsection{Validation of Proxy tasks}
\begin{figure}[h]
\begin{center}
 \includegraphics[width=0.9\linewidth]{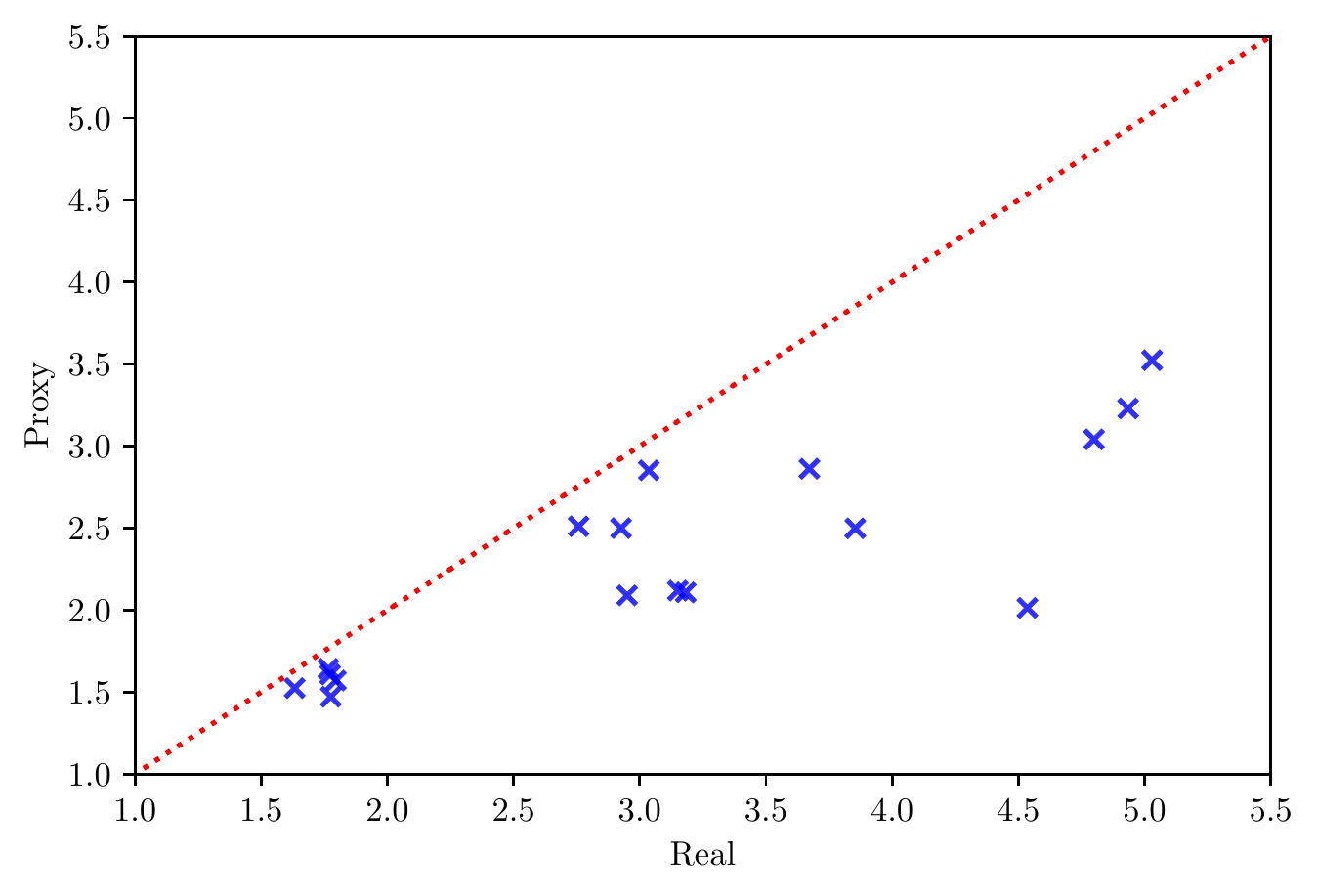}
\end{center}
\vspace{-1em}
  \caption{The correlation plot of real evaluation and proxy task performance on CIFAR-10.}
  \vspace{-1em}
\label{fig:correlation}
\end{figure}

The proxy task of our method is to directly evaluate IS on the child model of the shared GAN, which boosts the training speed dramatically. To validate the effectiveness of the designed proxy task, we train the derived architectures from scratch for 30 epochs and evaluate their IS, using AutoGAN on CFIAR-10.  We plot the correlation between the evaluation value provided by our proxy task (Proxy) and the true evaluation (Real) in Fig. \ref{fig:correlation}.  We can observe a positive correlation between each other, with a Spearman’s rank correlation coefficient of 0.779, demonstrating that the proxy task provides a fair approximation of the true evaluation.  We can also observe that the proxy evaluation tends to underestimate the Inception score, because of the uncompleted and shared training in our proxy task.

%demonstrating the effectiveness of our designed proxy task. %\textcolor{red}{A correlation figure will be plotted here to show the correlation of the true performance and the proxy task's evaluation results.}

%-------------------------------------------------------------------------
\subsubsection{Comparing to Using FID as Reward}
\begin{figure}[h]
\centering
    \begin{subfigure}[t]{0.238\textwidth}
        \centering
        \includegraphics[width=\textwidth]{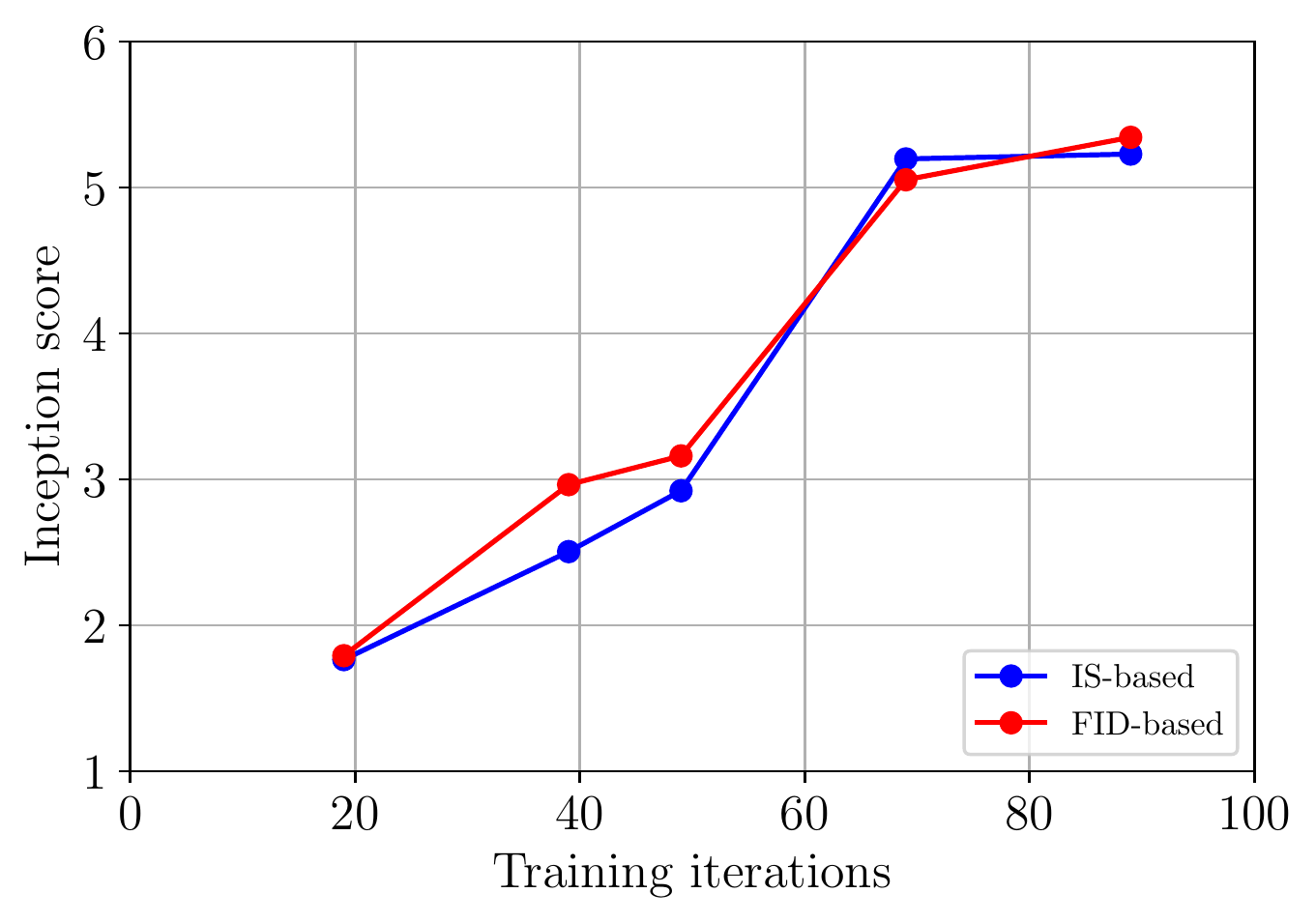}
        % \includegraphics[height=1.2in]{a}
        % \caption{Lorem ipsum}
    \end{subfigure}\hfill
    \begin{subfigure}[t]{0.238\textwidth}
        \centering
        \includegraphics[width=\textwidth]{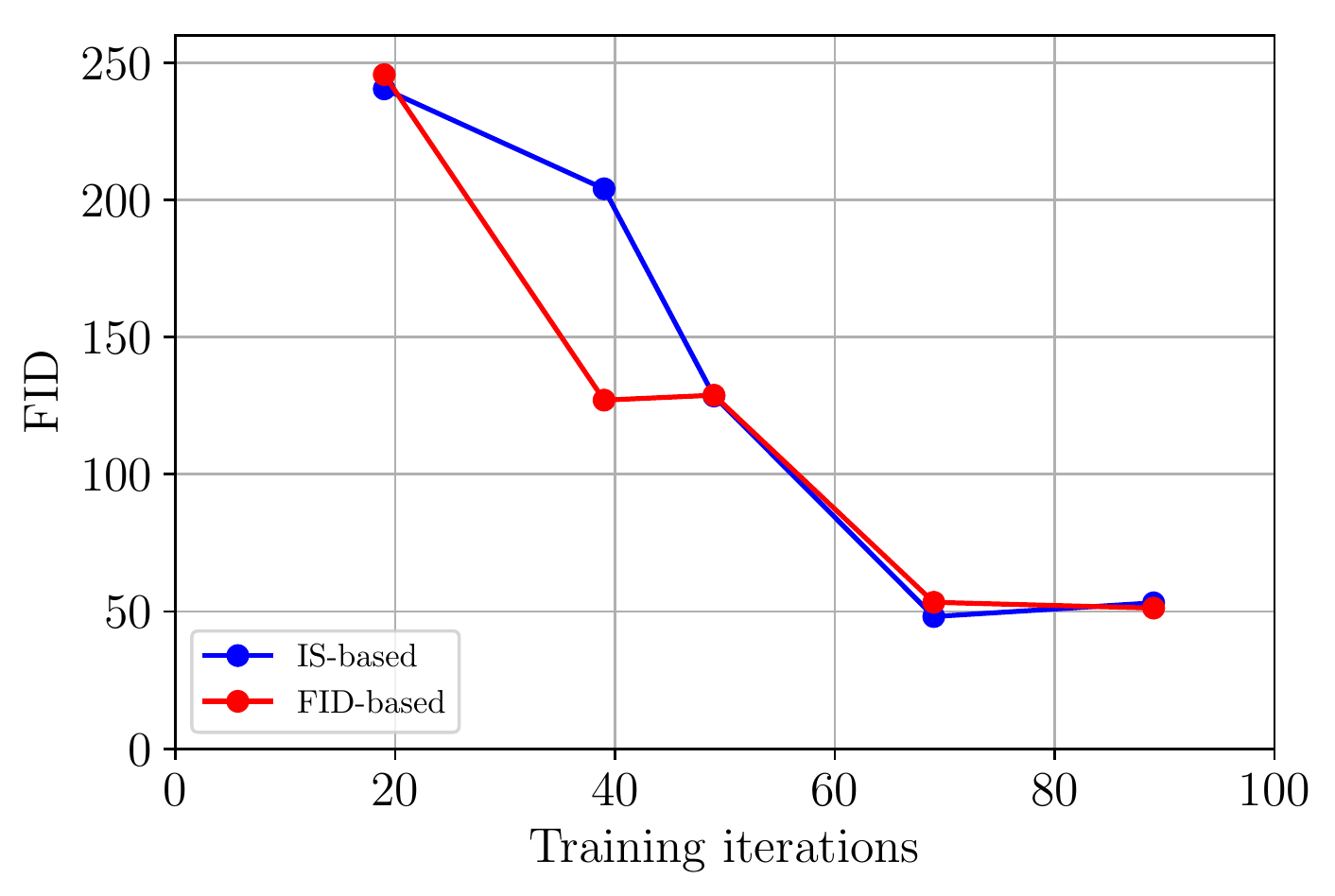}
        % \includegraphics[height=1.2in]{a}
        % \caption{Lorem ipsum}
    \end{subfigure}
  
  \caption{Comparison between using IS and FID (reciprocal) as the AutoGAN's reward. The model with IS-based reward is plotted in blue, while the one with FID-based reward in red. The top figure shows both models' IS values as training goes on, while the bottom presents FID values.
  %The curve of corresponding Inception score. We can see that using Inception score as reward outperforms using FID score as reward in on Inception score.
  }
  \vspace{-1em}
\label{fig:IS}
\end{figure}

Besides IS, another important evaluation metric for GAN is the FID score \cite{salimans2016improved}: the smaller FID indicates better generation quality. What if we use the reciprocal of FID, instead of IS, as the AutoGAN controller's reward?   To answer this question, on CIFAR-10, we search for another model from scratch, using the reciprocal of FID as the reward (everything else unchanged), and compare to the AutoGAN trained with IS reward.  During searching, for both models, we evaluate both IS and FID values periodically with the training progressing. Fig. \ref{fig:IS} plots the curves, from which we can see that while both rewards can increase as the search goes on, the FID-driven search (red curves) shows comparable performance with the IS-driven search (blue curves), under both metrics. However, computing FID is more time-consuming than calculating IS, as it requires to calculate the co-variance and mean of large matrices. Thus, we choose IS as the default reward for AutoGAN.

%where a smaller FID score indicates a better GAN model. Hence we try to the merge FID score into reward, where reward $R=c/FID$ (a small FID score indicated higher reward).

%-------------------------------------------------------------------------
\subsubsection{Parameter dynamic-resetting}
%In this section, we will discuss the effect of 

\begin{figure}[h]
\begin{center}
 \includegraphics[width=0.9\linewidth]{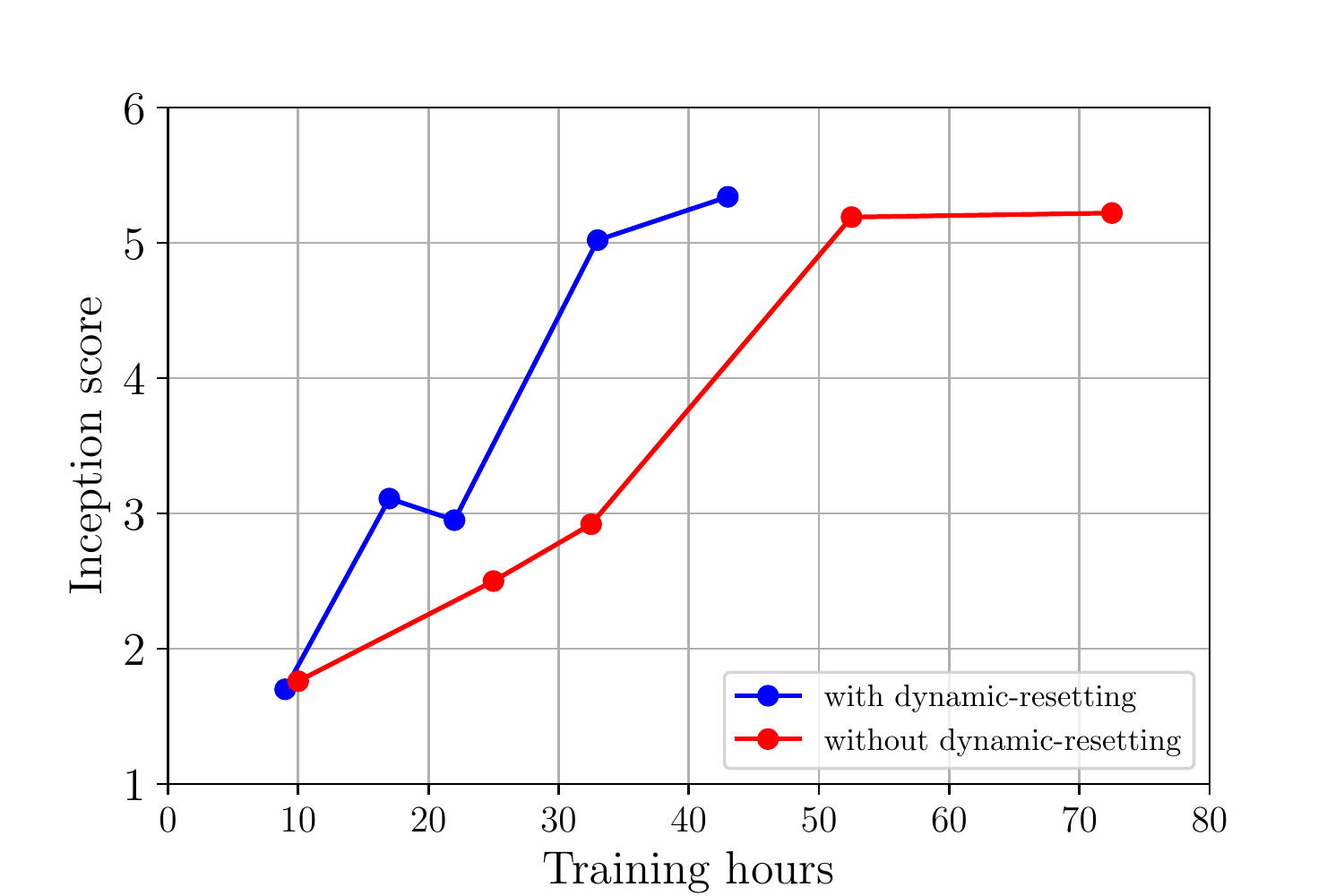}
\end{center}
\vspace{-1em}
   \caption{Comparing AutoGAN with (blue) and without (red) dynamic-resetting. Dynamic-resetting can boost training efficiency while achieving comparable performace.}
  % \vspace{-1em}
\label{fig:reset}
\end{figure}

It is well known that the training process of GAN is extremely unstable and prone to mode collapse \cite{brock2018large} after long time training. It could be a waste of time to continue training the shared collapsed model. Hence, we introduce parameter dynamic-resetting to alleviate the issue. For comparison, we conduct two AutoGAN experiments on CIFAR-10, with the parameter sharing strategy proposed in \cite{Pham:tl}, and our proposed dynamic-resetting plus parameter sharing strategy. We evaluate IS during both training processes. As plotted in Figure \ref{fig:reset}, we can see that they achieve comparable performance. However, the training process with dynamic-resetting only takes 43 hours, while it will take 72 hours without dynamic-resetting.

%-------------------------------------------------------------------------
\subsubsection{Multi-Level Architecture Search}
\begin{figure}[h]
\begin{center}
 \includegraphics[width=0.9\linewidth]{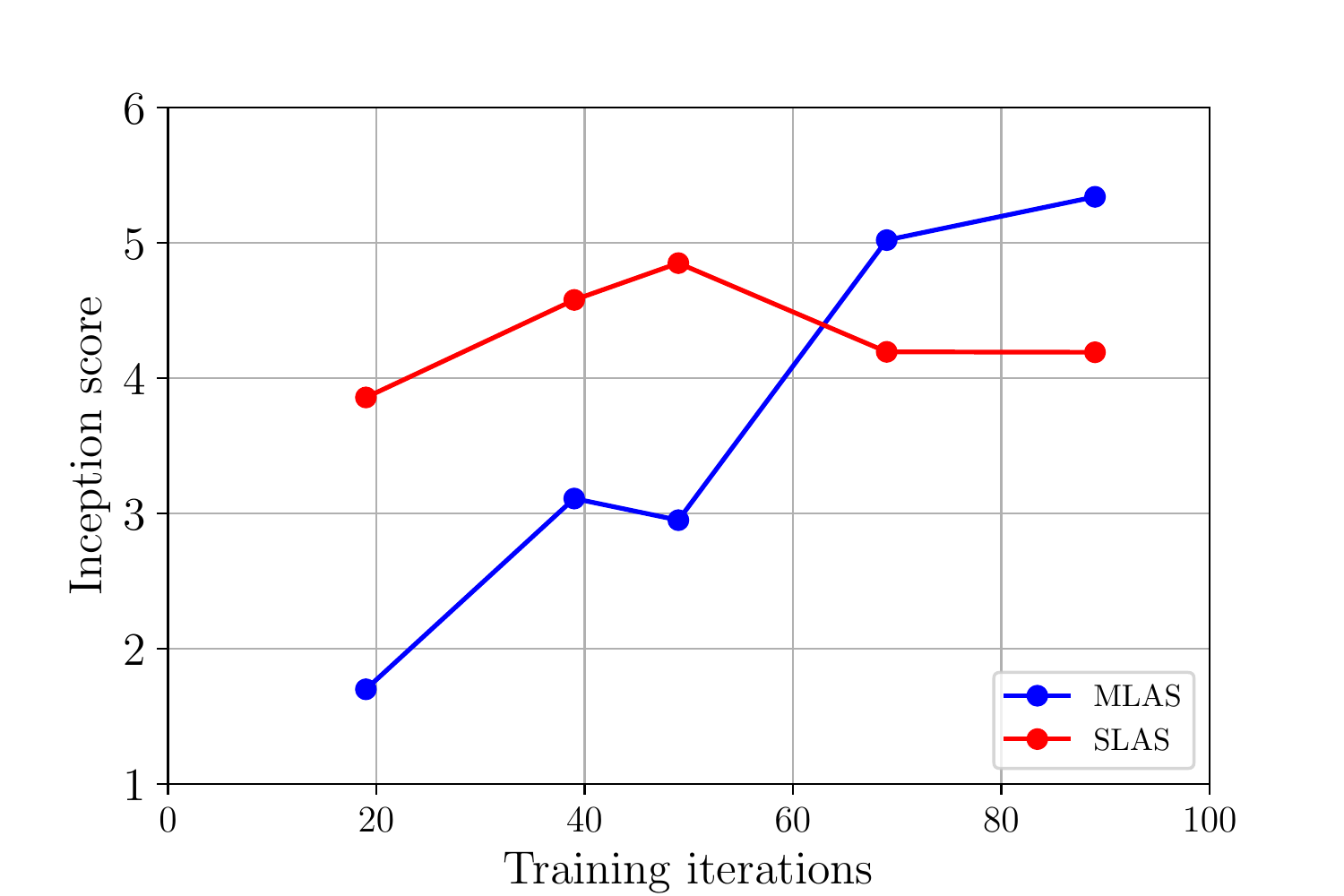}
\end{center}
\vspace{-1em}
   \caption{Comparison of AutoGAN with MLAS (blue line) and SLAS (red line) training chemes.}
   \vspace{-0.5em}
\label{fig:progressive}
\end{figure}

Our AutoGAN framework employs a multi-level architecture search (MLAS) strategy by default. For comparison, we conduct another AutoGAN experiment on CIFAR-10 with single-level architecture search (SLAS), where the whole architecture will be derived through a single controller at once. We evaluate ISs and compare with the training with MLAS. We can see that SLAS obtain high Inception score at the beginning, but the Inception score of MLAS grows progressively and finally outperforms SLAS. Besides, Training SLAS is much slower than training MLAS, as the generator output image will always be of the final resolution.  Figure \ref{fig:progressive} demonstrates the evident and consistent advantages of MLAS. 

\subsubsection{Comparison to random search}
We implemented the two random search algorithms in \cite{li2019random}: one with weight sharing and the other without weight sharing (early stopping). We re-searched AutoGAN with these two algorithms on CIFAR-10 and constrained the searching time for 48 hours. As a result, the discovered architecture with weight sharing achieves IS = 8.09 and FID = 17.34, while the other achieves IS = 7.97 and FID = 21.39. Both are inferior to our proposed search algorithm and endorse its effectiveness.

%-------------------------------------------------------------------------
\section{Conclusions, Limitations and Discussions}

AutoGAN presents the first effort on bringing NAS into GANs. It is able to identify highly effective architectures on both CIFAR-10 and STL-10 datasets, achieving competitive image generation results against current state-of-the-art, hand-crafted GAN models. The ablation study further reveals the benefit of each component. 

AutoGAN appears to be more challenging than NAS for image classification, due to the high instability and hyperparameter sensitivity of GAN training itself. Recall that at the initial stage of AutoML, it can only design small neural networks that perform on par with neural networks designed by human experts, and these results were constrained to small academic datasets such as CIFAR-10 and Penn Treebank \cite{zoph2016neural}. Likewise, despite its preliminary success and promise, there is undoubtedly a large room for AutoGAN to improve. 

In order to make AutoGAN further more competitive than state-of-the-art hand-designed GANs, we point out a few specific items that call for continuing efforts:
\begin{itemize}
% \vspace{-0.5em}
    \item The current search space of AutoGAN is limited, and some powerful GANs are excluded from the searchable range. It needs to be enlarged with more building blocks, that prove to be effective in the GAN literature. Referring to recent GAN benchmark studies \cite{lucic2018gans,kurach2018gan}, we consider extending our search space with the attention/self-attention \cite{zhang2018self}, the style-based generator \cite{karras2018style}, the relativistic discriminator \cite{jolicoeur2018relativistic} and various losses such as the Wasserstein loss \cite{arjovsky2017wasserstein}, among others. 

    \item We have not yet tested AutoGAN on higher-resolution image synthesis so far, e.g. ImageNet. While the same algorithm is directly applicable in principle, the computational cost would become prohibitively high. For example, the search on CIFAR-10 already takes 43 hours. The key challenge lies in how to further improve the search algorithm efficiency. 
    
    As one possible strategy, NAS in image classification typically applies transfer learning from low-resolution images to higher resolutions \cite{Zoph:2018ta}. It may be interesting, though challenging, to see how the similar idea will be applied to the image generation task of GANs, since it will be more demanding in preserving/synthesizing details than classification. 
%         \vspace{-0.3em}
    \item  We have not unleashed the potential of searching for better discriminators. We might formulate an alternating search between the generator and the discriminator, which can turn AutoGAN even far more challenging.
%                  \vspace{-0.3em}
    \item Eventually, AutoGAN will need the capability to incorporate labels, such as conditional GANs \cite{mirza2014conditional} and semi-supervised GANs \cite{salimans2016improved}. 
\end{itemize}

\clearpage

{\small
\bibliographystyle{ieee_fullname}
\bibliography{egbib}
}

\end{document}